\documentclass[oribibl]{llncs}
\usepackage{llncsdoc}
\usepackage{graphicx}
\usepackage{amssymb}
\usepackage{amsmath}
\usepackage{subfigure}
\usepackage{booktabs}
\usepackage{multirow}
\usepackage{url}

\urldef{\mailsa}\path|haoyou@ict.ac.cn|
\urldef{\mailsb}\path|lishiruilishirui@gmail.com|
\urldef{\mailsc}\path|{mohanlin,lihua}@ict.ac.cn|

\begin{document}
\title{Affine-Gradient Based Local Binary Pattern Descriptor for Texture Classification}

\author{You Hao\inst{1,2} \and Shirui Li\inst{1,2} \and Hanlin Mo\inst{1,2} \and Hua Li\inst{1,2}}

\institute{ 
	Key Laboratory of Intelligent Information Processing, 
	Institute of Computing Technology,
	Chinese Academy of Sciences,	Beijing, China
	\and
	 University of Chinese Academy of 
	 Sciences, Beijing, China\\
	\mailsa\\
	\mailsb\\
	\mailsc\\
	}

\maketitle
\begin{abstract}
	We present a novel Affine-Gradient based Local Binary Pattern (AGLBP) descriptor for texture classification. It is very hard to describe complicated texture using single type information, such as Local Binary Pattern (LBP), which just utilizes the sign information of the difference between pixel and its local neighbors. Our descriptor has three characteristics: 1) In order to make full use of the information contained in the texture,  the Affine-Gradient, which is different from Euclidean-Gradient and invariant to affine transformation, is incorporated into AGLBP. 2) An improved method is proposed for rotation invariance, which depends on the reference direction calculating respect to local neighbors. 3) Feature selection method, considering both the statistical frequency and the intraclass variance of the training dataset, is also applied to reduce the dimensionality of descriptors.   
	Experiments on three standard texture datasets, Outex12, Outex10 and KTH-TIPS2, are conducted to evaluate the performance of AGLBP. The results show that our proposed descriptor gets better performance comparing to some state-of-the-art rotation texture descriptors in texture classification.

\end{abstract}

\section{Introduction}
\label{sec:intro}
Texture is the most fundamental information on which the majority of all living organisms base their visual cognition and is a key component of  computer vision system\cite{haindl2013visual}. 
Basically, all the digital images can be regarded as texture. Texture analysis has been applied to many visual problems such as material categorization, surface inspection, medical image analysis, object recognition, image segmentation, pedestrian detection, face analysis and so on. 

Over the years, lots of texture descriptors have been proposed \cite{haralick1973textural,qian2009object,wu1996rotation,porter1997robust}. Among these descriptors, local patterns have achieved good performance in most texture applications \cite{ojala2002multiresolution,lowe2004distinctive,dalal2005histograms}. In particular, LBP is an efficient descriptor for describing local structures \cite{ojala2002multiresolution}.
LBP descriptors have already demonstrated powerful discriminative capability, low computational complexity, and low sensitivity to illumination variation. For further improving the discrimination of LBP, a large number of LBP variants have been proposed \cite{liu2017local}. Most of these changes make efforts on the following three directions.

First is to utilize different forms of information from the original textures. Guo et al. proposed Complete LBP which utilized the sign and magnitude information of local neighborhood in the descriptor \cite{guo2010completed}. Some other methods concentrate on the local derivative information respected to a local region, such as LDP \cite{zhang2010local}, CLDP \cite{yin2014multi}, LDDP \cite{guo2012local}, POEM \cite{vu2012enhanced} and so on.
Second is rotation invariance, which is an important topic in texture classification. Many methods have been proposed to achieve rotation invariance, such as SRP \cite{liu2012sorted,skibbe2012fast}, SIFT \cite{lowe2004distinctive} and so on.	
Third is feature selection. The exponential increasing in the number of features with the patch size is a limitation for the traditional LBP. The uniform LBP descriptor proposed by Ojala et al. \cite{ojala2002multiresolution} is the first attempt to solve this problem.

The main contributions of the paper are threefold.
Firstly, we propose the Affine-Gradient based method to describe texture information. Affine-Gradient (AG) has some properties that Euclidean-Gradient (EG) does not have, which will be elaborated detailedly in the following.
Secondly, an improved method for determining the local reference direction is proposed to reach rotation invariance, which is fast to compute and effective for the rotation transformations.
Finally, we propose a simple but effective feature selection method considering both the distribution of patterns and the intraclass variance on the training datasets. Experiments show that the proposed feature selection method not only increases the discriminative power but also reduce the dimension of descriptor effectively. 

\section{Affine-Gradient based Local Pattern Descriptor}
In this section we elaborate our approach in detail. First, we give a brief review of LBP. Second, we discuss how to make full use of multi-information, especially Affine-Gradient (AG), for texture classification. The properties of AG are discussed in detail. Then we discuss the method we proposed to achieve the rotation invariance. Finally, the criterion for feature selection are discussed.
\subsection{Overview of LBP Method}
Th traditional LBP operator extracts information that is invariant to local gray-scale variations in the image. It is computed at each pixel location, considering the values of a small circular neighborhood around the central pixel $q_c$.
Then, the LBP is defined as following:
\begin{equation} \label{equ:LBP} 
LBP_{R,P}=\sum_{p=0}^{P-1}s(g_p-g_c)\cdot2^p \quad\quad s(x)=\begin{cases}	
1, x\geq 0 \\
0, x<0
\end{cases}
\end{equation}
where $g_c$ is the central pixel and $g_p$ are the values of its neighbors. $p$ is the index of the neighbor, $R$ is the radius of the circular neighborhood and $P$ is the number of pixels in the neighborhood. Then the histogram of these patterns is used to describe the texture of the image. 

There are three obvious disadvantages of LBP. First, it has no rotation invariance.
Second, it is just 1-th order sign information used in the descriptor. Third is the exponentially length increasing with the parameter $R$. 
The proposed method has been improved in these three direction.

\subsection{Affine-Gradient based Descriptors}
In here, we propose the method based on the AG information to increase the discrimination of the descriptor.
The Euclidean Gradient (EG) can de defined as $G=\sqrt{I_x^2+I_y^2}$. It is 2-norm of gradient in Euclidean space that remains invariant only under Euclidean transformation. 

Olver et al. \cite{olver1999affine} proposed that there are two basic relative affine differential invariant of 2-order in two-dimensional affine spaces as following:
\begin{gather}
H=I_{xx}I_{yy}-I_{xy}^2\\
J=I_{xx}I_y^2-2I_xI_yI_{xy}+I_x^2I_{yy}
\end{gather}

All other 2-order differential invariants can be made up of these two expressions. And their ratios constitute absolute invariant of differential in affine space. The affine gradient magnitude ($affG$) can be defined as equation (\ref{equ:affG}). In order to avoid the calculation fault of zero-denominator, we can make some changes to the definition as $affG'$.

\begin{equation} \label{equ:affG} 
affG=\left|\frac{H}{J}\right|,\quad\quad affG'=\sqrt{\frac{H^2}{J^2+1}}
\end{equation}	

\begin{figure}[th]
	\centering
	\subfigure[]{ \label{fig:exampleofimage}
		\scalebox{0.14}{\includegraphics{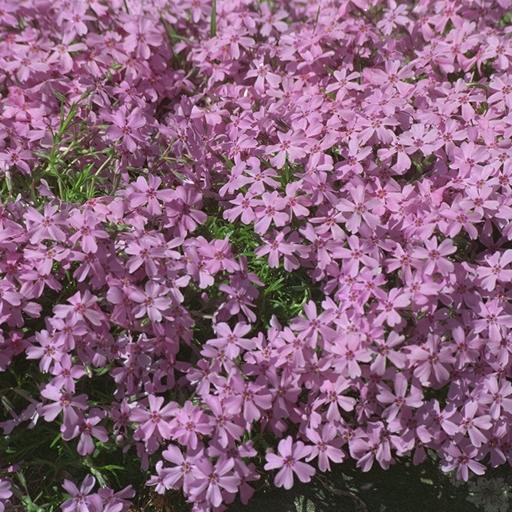}}}
	\subfigure[]{ \label{fig:g}
		\scalebox{0.14}{\includegraphics{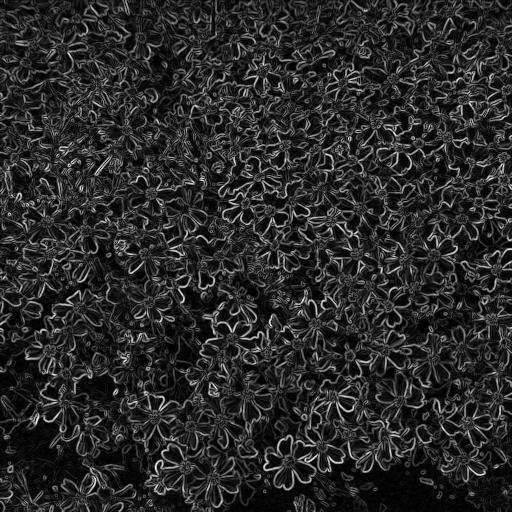}}}
	\subfigure[]{ \label{fig:affg1}
		\scalebox{0.14}{\includegraphics{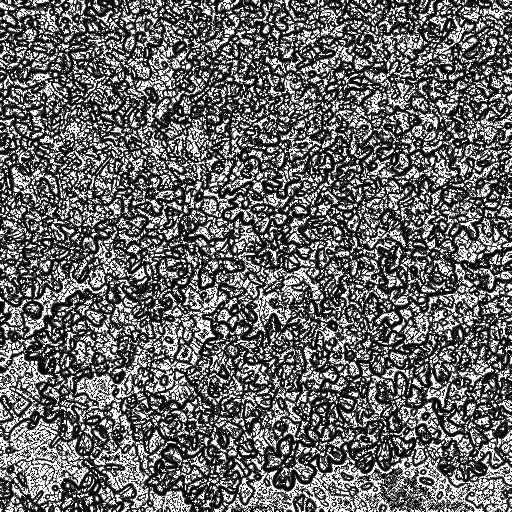}}}
	\subfigure[]{ \label{fig:affg2}
		\scalebox{0.14}{\includegraphics{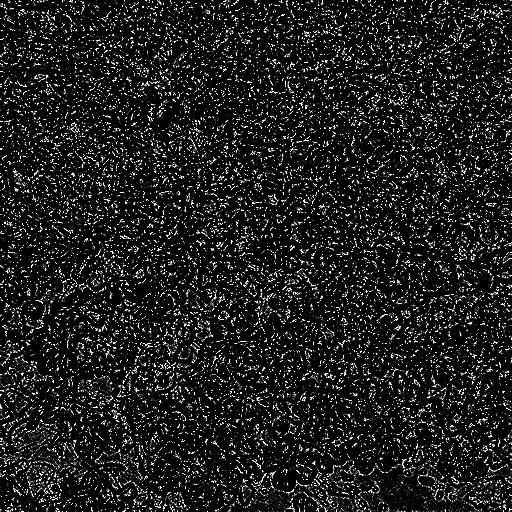}}}
	\caption{The EG and AG information of image example: (a) image example; (b) EG magnitudes of example; (c) AG of example range in (0-0.2); (d) AG of example range in (0.2-1).} \label{fig:gradient}
\end{figure}

The Affine-Gradient is superior than Euclidean-Gradient (EG), because AG is invariant for the affine transformation, and the EG just remains invariant under Euclidean transformation. Using the AG information can improve the robustness of descriptor for the geometric transformation. Ge et al. constructed a new descriptor using the AG to replace the EG in SIFT, which get much better performance than the original SIFT \cite{juan2013local}.  The gradient and AG information are shown in Fig. \ref{fig:gradient}.

\begin{figure}[th]
	\centering
	\subfigure[]{ \label{fig:histofgradient}
		\scalebox{0.3}{\includegraphics{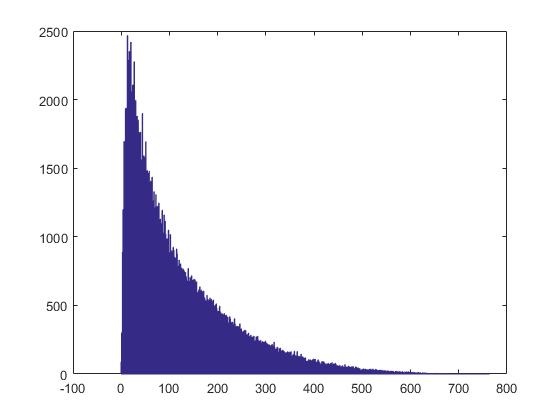}}}
	\subfigure[]{ \label{fig:histofaffg}
		\scalebox{0.3}{\includegraphics{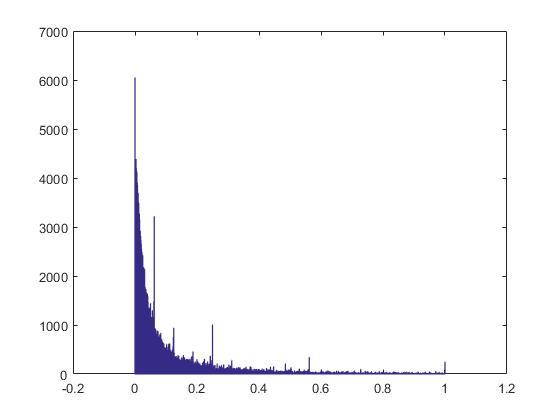}}}		
	\caption{The histogram of EG and AG: (a) histogram of the gradient; (b) histogram of the AG.} \label{fig:gradient2}
\end{figure}

In Fig. \ref{fig:histofgradient} and \ref{fig:histofaffg}, we can see that the histogram of EG is much more continuous and smooth than that of AG. In fact, the range of AG is from 0 to 162, not limited to 0 to 1 corresponding to Fig.\ref{fig:histofaffg}. It's just more sparse where the value bigger than 1. But the distribution of EG just ranges form 0 to 763 corresponding to Fig. \ref{fig:histofgradient}.
So intuitively, the information of AG ranging (0,1) probably corresponding to that of EG as shown in Fig. \ref{fig:g} and \ref{fig:affg1}. And there are some local extreme information in the AG as shown in Fig. \ref{fig:affg2}.

For further verification of the validity of AG, experiments are conducted on Outex12 dataset. The Local Gradient Pattern (LGP) and Local Affine-Gradient Patter (LAGP) can be defined as 
\begin{gather}
LGP_{R,P} = \sum_{p=0}^{P-1}s(G_p-G_c)\\
LAGP_{R,P} = \sum_{p=0}^{P-1}s(affG'_p-affG'_c)
\end{gather}
The $s$ function is defined in equation (\ref{equ:LBP}). The Multi-Information based descriptor MI-G,  can be defined as the concatenation of LGP and LBP. Similarly, MI-AG is the concatenation of LAGP and LBP. Then the experimental results are listed in Table \ref{table:expMI}.		

\begin{table}
	\caption{Results of Multi-Information based descriptors on Outex12}
	\label{table:expMI}
	\renewcommand{\arraystretch}{1.4}
	\setlength\tabcolsep{3pt}
	\begin{center}
	\begin{tabular}{lllll}
		\hline\noalign{\smallskip}
		Problem & form & $LBP$ & $MI\text{-}G$ & $MI\text{-}AG$ \\
		\noalign{\smallskip}
		\hline
		\noalign{\smallskip}
		\multirow{4}*{Outex12} 
		& $original$ & 55.26 & 58.04 & \textbf{58.69} \\
		& $ri$ & 71.37 & 73.49 & \textbf{79.28} \\  
		& $u2$ & 56.98 & 58.03 & \textbf{60.02} \\
		& $riu2$ & 65.09 &  77.62 & \textbf{77.65} \\
		\hline
	\end{tabular}
	\end{center}
\end{table}

From the results, we can see that the Multi-Information descriptor based on Affine-Gradient get the best performance in all scenarios.
It was demonstrated that the AG information can substantially increase the discriminative power of the descriptors.
\subsection{Rotation Invariance} 
Metha et al. \cite{mehta2016dominant} proposed a method that quantizing the directions into $P$ discrete values, then make direction with the maximum magnitude of the difference as the reference direction. But this definition discard the sign information of the magnitude and will assign the opposite directions into the same one.		
In this paper, we take both the sign and magnitude of the discrete directions into consideration. The reference direction can be defined as:
\begin{equation}
Ds = (\mathop{\arg \max}_{p\in(0,1,...,P-1)}{|g_p-g_c|} + \frac{P}{2} \cdot s(g_D-g_c))\mod P
\end{equation}		
where $s$ is the sign function defined in equation (\ref{equ:LBP}).
The proposed descriptor is computed by rotating the weights with respect to the reference direction. The rotation invariance LBP (roLBP) can be defined as

\begin{equation}
roLBP_{R,P} = \sum_{p=0}^{P-1}s(g_p-g_c)\cdot 2^{(p-Ds)\mod P}
\end{equation}			

Applying the reference direction selection method to the LAGP descriptor. We can get the rotation invariant descriptor roLAGP as following:

\begin{equation}
roLAGP_{R,P} = \sum_{p=0}^{P-1}s(affG'_p-affG'_c)\cdot 2^{(p-Ds)\mod P}
\end{equation}

Then the final descriptor AGLBP can be defined as the concatenation of roLBP and roLAGP.

\begin{equation}
AGLBP_{R,P} = roLBP_{R,P}\text{\_}roLAGP_{R,P}
\end{equation}
\subsection{Feature Selection}
It is observed the dimensionality of descriptors also increases exponentially with the number of neighboring pixels. In \cite{mehta2016dominant}, proposed a method depending on the distribution of patterns in the training dataset. Besides, some patterns may be negative to the final classification result. So in our method, the intraclass variance of training datasets is also chosen as the evaluation for feature selection.	

\begin{figure}
	\subfigure[ ]{ \label{fig:histofvar1}
		\scalebox{0.3}{\includegraphics{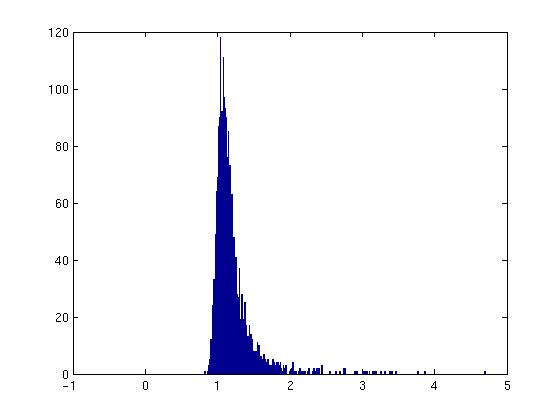}}}
	\subfigure[ ]{ \label{fig:histofvar2}
		\scalebox{0.3}{\includegraphics{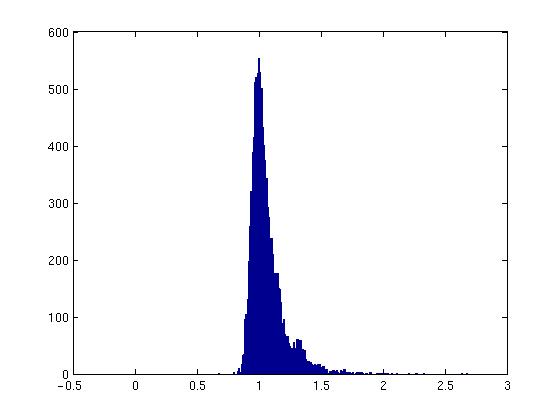}}}
	\caption{The intraclass variance distribution for roLBP on Outex12 dataset: (a) The variance distribution of roLBP in Outex12 training dataset; (b) The variance distribution of roLAGP in Outex12 training dataset.} \label{fig:histofvariance}
\end{figure}

In the statistical description, variance is defined as$\frac{1}{n-1}\sum(X-\mu)^2$, where $\mu$ is mean value of the array.	The distribution of the intraclass variance of all patterns are computed from the training dataset, as shown in Fig. \ref{fig:histofvariance}. 

\begin{figure}
	\subfigure[ ]{ \label{fig:curveofnum}
		\scalebox{0.3}{\includegraphics{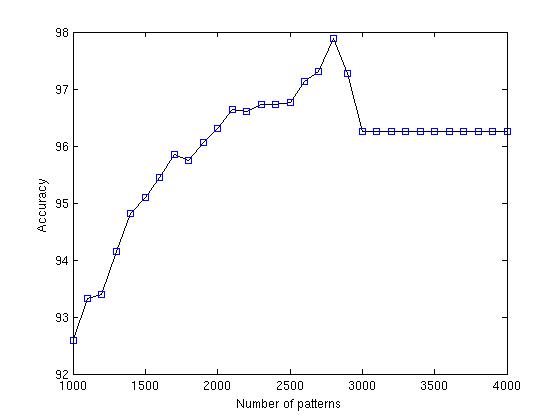}}}
	\subfigure[ ]{ \label{fig:curveofvar}
		\scalebox{0.3}{\includegraphics{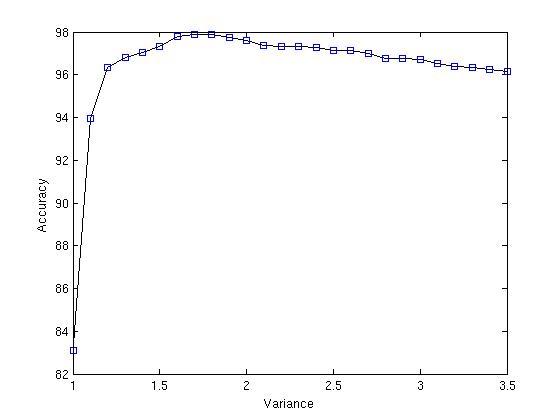}}}
	\caption{The accuracy-parameter curve for roLBP on Outex12 dataset: (a) the accuracy-N curve of roLBP on Outex12 dataset; (b) the accuracy-$\phi$ curve of roLBP on Outex12 dataset.} \label{fig:curve}
\end{figure}

The bins of the histogram are sorted in descending order. Then there will be two method for feature selection. One selects the top $N$ patterns in the ordered list, the other selects bins which is less than a threshold $\phi$ as the final descriptor. The final patterns selected depend on the threshold parameter $N$ or $\phi$ and the training datasets. The final dimensionality of the descriptor is not constant. It varies across different datasets.	
The accuracy-parameter curve of the two method for roLBP on Outex12 dataset are plotted in Fig. \ref{fig:curve}.

It can be observed in Fig. \ref{fig:curveofvar} that the classification accuracy reach the peak with the threshold value almost between 1.6-2.0, just over the peak of distribution corresponding to Fig. \ref{fig:histofvar1} . This values results in a significant reduction of the dimensionality. 

Thus, the proposed approach consider both the statical frequency and the intraclass variance of the training textures, which not only reduces the dimensionality of descriptors, but also improves the classification accuracy. The effective of the proposed approach will be demonstrated in next section.
\subsection{Classification method}
Some state-of-the-art methods, such as artificial neural network (ANN), SVM, AdaBoost, can achieve outstanding  classification performance, but these methods require complex learning procedure and may influence analysis of discriminative capabilities of features. To make a fair comparison with some other approaches, the Nearest Neighbor (NN) classifier based on the Chi-Square distance was performed as our classification method. The effectiveness of the Chi-Square distance for classification is demonstrated in \cite{guo2010descriptor,guo2012local}. 
\section{Experiments} \label{sec:exp}
To evaluate the proposed descriptor (AGLBP), three experiments are conducted on texture datasets: Outex10, Outex12 and KTH-TIPS2. Outex10 and Outex12 datasets are for rotation invariant texture classification with rotation and illumination deformations. The KTH-TIPS2 is for material categorization and includes scale and viewpoints variations. The parameter $\phi$ of proposed method is set to 2 in all our experiments.	
\subsection{Outex12}
Outex is a framework for empirical evaluation of texture classification algorithms\cite{ojala2002outex}.
First we conduct experiment on the Outex12 dataset. It consists of 9120 images, which are separated into 24 different texture classes captured with different illuminations and rotations. This dataset contains 20 training images and 360 (2*9*20) testing images under two different illumination and 9 different orientation for each class. In experiment, following two problem proposed in the dataset\cite{ojala2002outex}, problem 000 and 001. Considering the length of the final descriptor is depending on the parameter (R,P), we use a conservative setting of the parameter as (1,8),(2,12),(3,16). All the LBP-based methods were performed and the results are shown in Table \ref{table:exp1}. 

\begin{table}	
	\label{table:exp1}
	\caption{Experiment results of LBP based methods on different datasets}
	\renewcommand{\arraystretch}{1.4}
	\setlength\tabcolsep{3pt}
	\begin{center}
		\begin{tabular}{llccccccccc}
			\hline\noalign{\smallskip}
			Problems & (R,P) & $LBP$ & $LBP^{u2}$ & $LBP^{ri}$ & $LBP^{riu2}$ & $LBP\text{-}HF$ & $LBPV$ & $AGLBP$ \\
			\noalign{\smallskip}
			\hline
			\noalign{\smallskip}
			\multirow{3}*{Outex10} 
			& (1,8) 	& 50.20 & 57.44 & 82.78 & 74.38 & 72.03	& 91.40 & 63.72\\  
			& (2,12)    & -		& 59.62 & 91.48 & 86.74	& 90.52	& 92.18 & \textbf{95.43}\\
			& (3,16) 	& - 	& 61.35 & 95.76 & 88.92 & 97.03	& 94.37 & \textbf{99.22}\\
			\noalign{\smallskip}
			\hline
			\noalign{\smallskip}
			\multirow{3}*{Outex12-000} 
			& (1,8)		& 54.21 & 55.81 & 72.26 & 65.93 & 70.85 & 76.41 & 61.99\\  
			& (2,12)	& -		& 57.85 & 86.78 & 82.66 & 88.49 & 86.80 & \textbf{93.31}\\
			& (3,16)	& -		& 58.56 & 93.50 & 83.98 & 91.08 & 90.85 & \textbf{97.84}\\
			\noalign{\smallskip}
			\hline
			\noalign{\smallskip}
			\multirow{3}*{Outex12-001} 
			& (1,8)		& 56.32 & 58.15 & 70.39 & 64.26 & 77.24 & 77.08 & 67.50\\  
			& (2,12)	& -		& 57.08 & 84.77 & 75.86 & 91.34 & 84.09 & \textbf{94.83}\\
			& (3,16)	& -		& 59.49 & 92.97 & 79.63 & 92.40 & 84.76 & \textbf{97.38}\\
			\noalign{\smallskip}
			\hline
			\noalign{\smallskip}
			\multirow{4}*{KTH-TIPS2} 
			& (1,8)		& 90.97 & 85.85 & 83.65 & 82.78 & 88.73 & 78.98 & 81.28\\  
			& (2,12)	& -		& 87.92 & 89.75 & 87.95 & 90.87 & 83.00 & \textbf{95.23}\\
			& (3,16)	& -		& 91.95 & 94.36 & 91.52 & 91.85 & 85.10 & \textbf{97.12}\\
			\hline
		\end{tabular}
	\end{center}	
\end{table}

Among these methods, the proposed method with setting (3,16) has achieved the highest accuracy of 97.84\% for problem 000 and 97.38\% for problem 001. For further analysis, we compare our method with some other state-of-the-art methods. The results are shown in Table \ref{table:exp2}. It can be seen that the proposed descriptor achieves the best result, the close second is $DRLBP$, which get the accuracy 97.15\% for problem 000 and 95.37\% for problem 001.

\subsection{Outex10}
Then experiment is conducted on the Outex10 dataset, which includes 4320 images of 24 different classes. These images are captured under the same illumination but  rotated at nine different angles. There are 20 images at each angle for each class. Following the problem proposed in the dataset\cite{ojala2002outex}, 480 images captured at angle $0^\circ$ are taken as the training set and the rest 3840 images captured at other angles used for testing.	

The results with various setting are shown in Table \ref{table:exp1}. For further analysis, AGLBP are compared with some other state-of-the-art approaches. The result of these methods are also shown in Table \ref{table:exp2}. It can be observed that AGLBP performs well under various rotation deformations. Among all, our method with setting (3,16) has achieved the highest accuracy 99.22\%, just a little improvement on the results to the 99.19\%, which achieved by $DRLBP$.

\subsection{KTH-TIPS2 Dataset}	

Experiment on the KTH-TIPS2 dataset has also been conducted for material classification. The KTH-TIPS2 database contains 11  texture classes with different materials. For each class, the images are captured from 4 different samples of materials. And for each sample, 9 different scales with 4 different illumination and 3 different poses are conducted for the imaging. In this experiment, following problem proposed in most research\cite{guo2010rotation,guo2011texture}, images of one random sample are selected from each class are taken as the training dataset, images from the other samples are taken as the testing dataset.

All the methods were performed and the results are shown in Table \ref{table:exp1}. As the same, AGLBP is also compared with some other state-of-the-art approaches. The result of these methods are shown in Table \ref{table:exp2}. The proposed descriptor outperforms all other descriptors again. It can be concluded that our method is effective for texture classification.
\begin{table}
	\label{table:exp2}
	\caption{Experiment results of descriptors on different datasets}	
	\renewcommand{\arraystretch}{1.4}
	\setlength\tabcolsep{3pt}
	\begin{center}	
		\begin{tabular}{lcccccccccc}
			\hline\noalign{\smallskip}
			Problems 	& $LBP^{ri}$ & $LDDP$ & $LCP$ & $LBP\text{-}HF$ & $LBPV$ & $VZ\_MR8$ &$VZ\_Joint$    \\
			\noalign{\smallskip}
			\hline
			\noalign{\smallskip}
			Outex10		&  95.76 	& 73.16 & 74.12 & 97.03 & 94.37 & 93.59	& 92.00 \\
			Outex12-000	& 93.50		& 63.48 & 70.16 & 91.08 & 90.85 & 91.34 & 90.46 \\
			Outex12-001	& 92.97		& 68.48 & 68.48 & 92.40 & 84.76 & 92.83 & 91.74 \\
			KTH-TIPS2	& 94.36		& 92.74 & 92.15 & 91.85 & 85.10 & 93.50 & 95.46  \\
			\noalign{\smallskip}
			\hline 	
			\noalign{\smallskip}
			Problems 	& $PLBP$ 	& $MDLBP$ & $FBLLBP$  & $BIF$ & $LEP$  & $DRLBP$ & $AGLBP$ \\
			\noalign{\smallskip}
			\hline
			\noalign{\smallskip}
			Outex10 	& 96.64 	& 95.34 	& 98.68 & -		& -		& 99.19 & \textbf{99.22}\\
			Outex12-000	& 82.79 	& 93.96		& 88.38 & -		& -		& 97.15 & \textbf{97.84}\\
			Outex12-001	& 90.08 	& 89.94		& 92.17 & -		& -		& 95.37 & \textbf{97.38}\\
			KTH-TIPS2 	& - 		& -			& -		& 98.50 & 96.41 & 96.78 & \textbf{97.12}\\
			\hline
		\end{tabular}
	\end{center}	
\end{table}	

\section{Conclusion}
In this paper we have proposed an Affine-Gradient based Local Binary Pattern (AGLBP) descriptor for texture classification. Affine-Gradient is different from the Euclidean-Gradient and has been proved to have a good improvement for texture classification. In addition, we have proposed an improved method for determining the local reference direction to reach rotation invariance. Importantly, the dimension increasing bringing by multi-information is also alleviated by proposed feature selection method, which considering both the statistical frequency and the intraclass variance of the training texture. Three extensive experiments have been conducted on texture datasets including rotating, scaling and viewpoint deformations. The results demonstrate that the AGLBP performed better than some state-of-the-art approaches for texture classification.	The AGLBP utilize the Affine-Gradient which has been demonstrated robust for the viewpoint deformation. For further research, information invariant for projective transformation should be utilized to enhance the robustness to viewpoint deformation.

\bibliographystyle{splncs03}
\bibliography{egbib}
\end{document}